\documentclass[sigconf]{acmart}
\pdfoutput=1
\usepackage{booktabs} 
\usepackage[flushmargin, multiple]{footmisc}
\usepackage{multirow}
\usepackage{subfig}
\usepackage{CJKutf8}

\newcommand{\figref}[1]{Figure\hspace{1mm}\ref{#1}}
\newcommand{\tabref}[1]{Table\hspace{1mm}\ref{#1}}

\def\BibTeX{{\rm B\kern-.05em{\sc i\kern-.025em b}\kern-.08emT\kern-.1667em\lower.7ex\hbox{E}\kern-.125emX}}
    
\copyrightyear{2019}
\acmYear{2019} 
\setcopyright{acmcopyright}
\acmConference[KDD '19]{The 25th ACM SIGKDD Conference on Knowledge Discovery and Data Mining}{August 4--8, 2019}{Anchorage, AK, USA}
\acmBooktitle{The 25th ACM SIGKDD Conference on Knowledge Discovery and Data Mining (KDD '19), August 4--8, 2019, Anchorage, AK, USA}
\acmPrice{15.00}
\acmDOI{10.1145/3292500.3330789}
\acmISBN{978-1-4503-6201-6/19/08}

\begin{document}

\title[Support the Creation of Effective Ad Creatives]{Conversion Prediction Using Multi-task Conditional Attention Networks to Support the Creation of Effective Ad Creatives}

\author{Shunsuke Kitada}
\affiliation{%
  \institution{Hosei University}
  \city{Tokyo}
  \country{Japan}
}
\email{shunsuke.kitada.8y@stu.hosei.ac.jp}

\author{Hitoshi Iyatomi}
\affiliation{%
  \institution{Hosei University}
  \city{Tokyo}
  \country{Japan}
}
\email{iyatomi@hosei.ac.jp}

\author{Yoshifumi Seki}
\affiliation{%
  \institution{Gunosy Inc}
  \city{Tokyo}
  \country{Japan}
}
\email{yoshifumi.seki@gunosy.com}

\renewcommand{\shortauthors}{S. Kitada et al.}

\begin{abstract}
Accurately predicting conversions in advertisements is generally a challenging task, because such conversions do not occur frequently.
In this paper, we propose a new framework to support creating high-performing ad creatives, including the accurate prediction of ad creative text conversions before delivering to the consumer.
The proposed framework includes three key ideas: multi-task learning, conditional attention, and attention highlighting.
Multi-task learning is an idea for improving the prediction accuracy of conversion, which predicts clicks and conversions simultaneously, to solve the difficulty of data imbalance.
Furthermore, conditional attention focuses attention of each ad creative with the consideration of its genre and target gender, thus improving conversion prediction accuracy.
Attention highlighting visualizes important words and/or phrases based on conditional attention.
We evaluated the proposed framework with actual delivery history data (14,000 creatives displayed more than a certain number of times from Gunosy Inc.), and confirmed that these ideas improve the prediction performance of conversions, and visualize noteworthy words according to the creatives' attributes.
\end{abstract}

%
%



\maketitle

\section{Introduction}
\renewcommand{\thefootnote}{\fnsymbol{footnote}}
\footnotetext[1]{This work was conducted while the first author was doing an internship at Gunosy Inc. We thank the ad engineering team who provided useful comments.}
\renewcommand{\thefootnote}{\arabic{footnote}}

\begin{figure}[tb]
    \centering
    \includegraphics[width=0.8\linewidth]{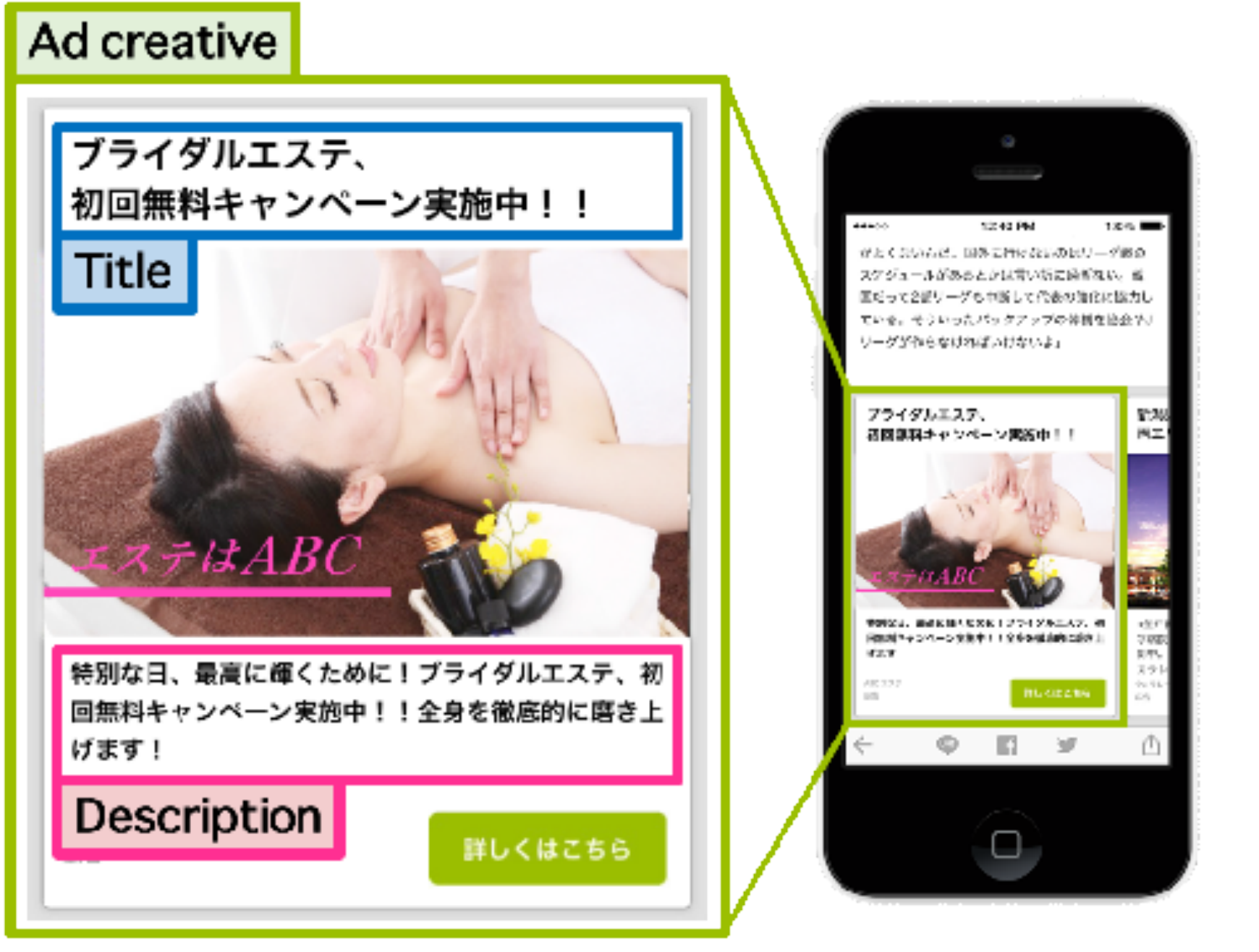}
    \caption{
        An example of an ad creative in digital advertising: an ad creative is constructed with an image and two short texts. These short texts are called the title and the description.
    }
    \label{fig:ad_creative_sample}
\end{figure}

In display advertisements, ad creatives, such as images and texts, play an important role in delivering product information to customers efficiently~\cite{chapelle2015simple}. 
\figref{fig:ad_creative_sample} shows an example of an ad creative which is constructed by two short texts and an image.
The performance of these advertisements is generally defined by \textit{the revenue of conversions} per \textit{the cost of the advertisement}.
Conversions are user actions, such as the purchase of an item or the download of an application, and they represent a known metric that advertisers try to maximize through their ad creatives. 
The costs of advertisements are generally calculated by the cost per click (CPC), where an advertiser pays for the number of times their advertisement has been clicked.
Therefore, the high performance of an ad is determined by minimizing the amount paid for the maximum number of conversions.
Creating high-performing ad creatives is a difficult but crucial task for advertisers. 

The purpose of this study is to supporting the creation of ad creatives with many conversions, and we propose a new framework to support creating high-performing ad creatives, including accurate prediction of ad creative text conversions before delivery to the consumer\footnote{We have also improved the CVR prediction using the result of conversion prediction.}.
If conversions of ad creatives can be predicted before delivery to consumers, advertisers can avoid the losses incurred by the high cost of ineffective advertisements.
Moreover, because ad creatives with high click-through rates (CTRs), and low conversions have a tendency to deceive users, we also expect to improve the user experience on media displaying those ads. 
As a result, advertisers will be able to focus on improving the CTR of ad creatives.

Some attempts to support the creation of high-performing creatives by predicting ad creative conversions have been reported in the industry\footnote{\url{https://www.facebook.com/business/m/facebook-dynamic-creative-ads}}\footnote{\url{https://www.adobe.com/en/advertising/creative-management.html}}\footnote{\url{https://support.google.com/google-ads/answer/2404190?hl=en}}, but as far as we know, no academic research has been published in this area.
Thomaidou et al.~\cite{thomaidou2014toward, thomaidou2013grammads} proposed a framework for generating ad creatives automatically.
However, this framework focuses on search ads, and generates ad text according to set rules.
Thus, this framework cannot be applied for our purpose.
Some studies have reported that ad creatives affect the CTR of advertisements~\cite{azimi2012visual,cheng2012multimedia, bruce2017dynamic}, but they do not predict the conversions.
Prediction of a user's CTR or conversion rate (CVR) is a general task undertaken by many studies in this research area, but there are no studies that have predicted these rates for ad creatives.
The prediction of an ad creative's performance is another important issue, but to the best of our knowledge, no study has examined this issue.

Although ad creatives are mainly image and text, we focus on the latter, and predicting its conversions.
Because it is difficult to replace ad images, but easy to replace text, in this work, we propose a recurrent neural network (RNN)-based framework that predicts the performance of an ad creative text before delivery.
The proposed framework includes three key ideas, namely, multi-task learning, conditional attention, and attention highlighting.
Multi-task learning is an idea for improving the prediction accuracy of conversion, which predicts clicks and conversions simultaneously, to solve the difficulty of data imbalance.
Conditional attention focuses on the feature representation of each creative based on its genre and target gender, thus improving conversion prediction accuracy.
Attention highlighting visualizes important words and/or phrases based on conditional attention.
We confirm that the proposed framework outperforms some baselines, and the proposed ideas are valid for conversion prediction.
These ideas are expected to be useful for supporting the creation of ad creatives.

This research is motivated to support the creation of high performing creative text. 
The contributions are summarized as follows:
\begin{enumerate}
    \item We propose a new framework that accurately predicts ad creative performance.
    
    \vspace{1.5mm}
    To realize this, we propose two key strategies to improve the prediction performance of advertisement conversion.

    \vspace{2mm}
    \begin{enumerate}
        \item Multi-task learning predicts conversion, together with previous click actions, by learning common feature representations. 
        \item The Conditional attention mechanism focuses attention on the feature representation of each creative text considering the target gender and genre.
    \end{enumerate}
    
    \vspace{2mm}
    \item We propose attention highlighting that offers important words and/or phrases using conditional attention.
\end{enumerate}

A prototype implementation of the proposed framework with Chainer~\cite{chainer_learningsys2015} has been released on GitHub\footnote{\url{https://github.com/shunk031/Multi-task-Conditional-Attention-Networks}}.

\section{Related Work}\label{sec:related}
This study focuses on ad creatives.
First, we describe existing studies that analyze high-performing ad creatives, and discuss how to generate them.
Many studies on advertising creatives focus on images, and offer few results for texts.
Furthermore, these studies focus on the CTR, rather than conversions.
Second, we introduce studies on performance prediction for ads.
In contrast to this study, which aims to predict the performance of new ads, these studies focus on images.
Finally, highlighting studies related to our ideas, we introduce multi-task learning and RNN-based attention mechanisms. 
\vspace{-2.5mm}
\vspace{-2.5mm}

\subsection{Analysis and Generation of Effective Advertisements}
Because ad creatives play an important role in the performance of ads, some studies analyzed ad creative performance~\cite{bruce2017dynamic, azimi2012visual, cheng2012multimedia}.
For example, Azimi et al. ~\cite{azimi2012visual} tried to predict some features of the CTR using ad creative images, and evaluated the effectiveness of visual features.
The motivation of their study is similar to ours, but we focus on text instead of images in ad creatives and predict conversions rather than the CTR.
Cheng et al.~\cite{cheng2012multimedia} proposed a model for predicting the CTR of new ads, and reported some knowledge using feature importance, but the text features of that study were based on fixed rules.
With the development of deep learning, especially convolutional neural networks (CNNs)~\cite{krizhevsky2012imagenet}, visual features can be easily and effectively used for machine learning.
Chen et al.~\cite{chen2016deep} proposed Deep CTR, showing that using the features of ad images can significantly improve CTR prediction.

Thomaidou et al.~\cite{thomaidou2013grammads} developed GrammaAds, which automatically generates keywords for search ads.
In addition, they proposed an integrated framework for the automated development and optimization of search ads~\cite{thomaidou2014toward}.
These studies support the creation of text ad creatives, but because these methods are rule-based, focusing only on search ads, the methods cannot be applied to display advertising.
\vspace{-1.5mm}

\subsection{CTR and Conversion Prediction in Display Advertising}
CTR prediction of display advertising is important not only in the industry but also in academia.
In \cite{chakrabarti2008contextual,richardson2007predicting}, a CTR prediction model was proposed using logistic regression (LR), and factorization machines (FMs) have been proposed to predict advertising performance~\cite{rendle2010factorization, juan2016field, juan2017field}. 
In industry, LR and FMs are mainly used, because in display advertising, the prediction response time needs to be short to display an advertisement smoothly.
In recent years, deep neural networks (DNNs) have been applied for predicting the advertisement CTR~\cite{covington2016deep,cheng2016wide, guo2017deepfm, lian2018xdeepfm, chen2016deep}, 
and especially, some models combining DNNs with FMs have been proposed, and have improved  predictions~\cite{cheng2016wide, guo2017deepfm, lian2018xdeepfm, liu2018field}.
The improvements achieved by these models show that explicit interaction between variables is important for advertisement performance prediction, so we adopted explicit interaction in our idea as a conditional attention mechanism.

There are several studies on CVR prediction~\cite{punjobi2018robust, yang2016large, lu2017practical}, but there are not as many as the studies on CTR prediction.
CVR prediction is difficult, because the number of conversions is imbalanced data that almost ad creative's conversions are zero.
Existing studies tackled this difficulty.
Yang et al.~\cite{yang2016hierarchical} adopted dynamic transfer learning for predicting the CVR, and demonstrating feature importance.
Punjobi et al.~\cite{punjobi2018robust} proposed robust FMs for overcoming user response noise.
In this study, we tackle this difficulty using multi-task learning.
\vspace{-1.5mm}

\subsection{Background of the Proposed Strategies}
In this paper, we propose two key strategies for improving the prediction performance of advertisement conversion, namely, multi-task learning and a conditional attention mechanism.
As the background of these strategies, we describe multi-task learning and the RNN-based attention mechanism.

\textbf{Multi-task Learning}.
Multi-task learning~\cite{caruana1997multitask} is a method that involves learning multiple related tasks. It improves the prediction performance by learning common feature representations.
Recently, multi-task learning has been used in various research areas, especially natural language processing (NLP)~\cite{collobert2011natural,luong2015multi} and computer vision~\cite{zhang2014facial, liu2015multi, chu2015multi}, and has achieved significant improvements.
Conversions represent extremely imbalanced data, so conversion prediction is difficult.
Because ad click actions represent a pre-action of conversion actions, click prediction may be related to conversion prediction.
Therefore, we adopt multi-task learning, which predicts clicks and conversions simultaneously.

\textbf{RNN-based Attention Mechanism}.
For supporting the creation of ad creative text, we use the knowledge of NLP.
RNN-based models, such as long short-term memory (LSTM)~\cite{hochreiter1997long}, gated recurrent unit (GRU)~\cite{chung2014empirical}, and attention mechanisms~\cite{bahdanau2014neural} have made breakthroughs in various NLP tasks, for example, machine translation~\cite{bahdanau2014neural}, document classification~\cite{yang2016hierarchical, lin2017structured}, and image captioning~\cite{xu2015show}.
An RNN is a deep learning model for learning sequential data, and in NLP, this model can learn word order.
Attention mechanisms compute an alignment score between two sources, and make significant improvements in some NLP tasks.
Recently, self-attention~\cite{lin2017structured}, which computes alignment in a single source, was proposed.
In addition, visual analysis using attention can highlight important phrases and/or words using the attention result, so the attention mechanism is also attractive for interpretability.
In this study, we adopt a self-attention mechanism for improving conversion prediction performance and visualizing word importance.
    
\section{Methodology}\label{sec:method}
\begin{figure}[tb]
    \centering
    \includegraphics[width=\linewidth]{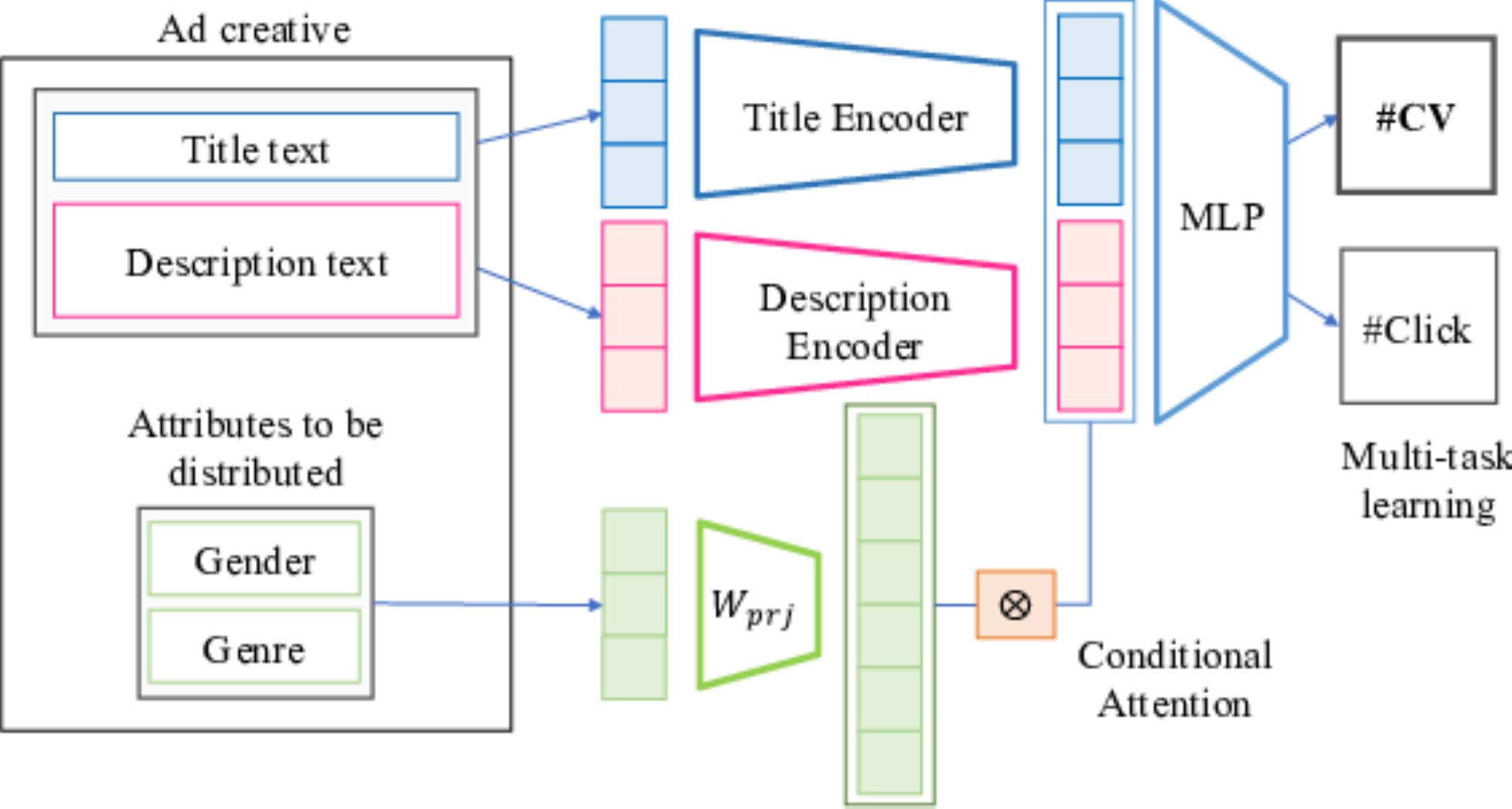}
    \caption{
        Outline of the proposed framework.
        In the framework, we propose two strategies: multi-task learning, which simultaneously predicts conversions and clicks, and a conditional attention mechanism, which detects important representations in ad creative text according to the text's attributes.
    }
    \label{fig:proposed_architecture}
\end{figure}

The outline of the proposed framework for evaluating ad creatives is shown in \figref{fig:proposed_architecture}.
In the framework, we propose two strategies: multi-task learning, which simultaneously predicts conversions and clicks, and a conditional attention mechanism, which detects important representations in ad creative text according to the text's attributes.

Conversion prediction using ad creatives with an imbalanced number of conversions is a challenging task.
Therefore, in multi-task learning, we expect to improve the model accuracy by predicting conversions along with clicks.
The conditional attention mechanism makes it possible to dynamically compute attention according to the attributes of the ad creatives, its genre, and the target gender.

\subsection{Framework Overview}

The input of the proposed framework is ad creative text and ad creative attribute values.
\figref{fig:ad_creative_sample} shows an example of an ad creative, and these are two short texts which are called titles and descriptions.
The ad attribute values are the gender of the delivery target and the genre of the ad creative, and they are related to the ad creatives.

Specifically, the input of the proposed framework is an ad creative text $S = \{\mathbf{w}_1, \mathbf{w}_2, \cdots, \mathbf{w}_n\}$ consisting of $n$ word embeddings,
where $\textbf{w}_{i} \in \mathbb{R}^{d_{w}}$ represents the word vector at the $i$-th position in the ad creative text.
Therefore, $S \in \mathbb{R}^{n \times d_w}$ is a two-dimensional matrix of the word sequence.

Incidentally, in the practical situation, a number of ad creative texts that have title and description texts are created for the target product.
These texts often have different contexts for maximizing the amount of information empirically.
Therefore, the proposed framework uses two \textit{text encoders}, which learn the individual context from the title and the description.

As a \textit{text encoder}, we adopted the GRU, which can extract features from ad creative text considering word order.
Specifically, title text $S^{\textrm{title}} = \{\mathbf{w}^{\textrm{title}}_{1}, \mathbf{w}^{\textrm{title}}_{2}, \cdots, \mathbf{w}^{\textrm{title}}_{n}\}$ and description text $S^{\textrm{desc}} = \{\mathbf{w}^{\textrm{desc}}_{1}, \mathbf{w}^{\textrm{desc}}_{2}, \cdots, \mathbf{w}^{\textrm{desc}}_{n}\}$ are input from the ad creative into title and description encoders,  respectively, and are encoded into feature representations as $\mathbf{h}^{\textrm{title}}_{t} \in \mathbb{R}^{u_{\textrm{title}}}$ and $\mathbf{h}^{\textrm{desc}}_{t} \in \mathbb{R}^{u_{\textrm{desc}}}$; $t=1, 2, \cdots, n$:
\begin{eqnarray}
    \begin{gathered}
        \mathbf{h}^{\textrm{title}}_{t} = \textrm{title~encoder}(\mathbf{w}^{\textrm{title}}_{t}, \mathbf{h}^{\textrm{title}}_{t-1}), \\
        \mathbf{h}^{\textrm{desc}}_{t} = \textrm{description~encoder}(\mathbf{w}^{\textrm{desc}}_{t}, \mathbf{h}^{\textrm{desc}}_{t-1}).
    \end{gathered}
\end{eqnarray}

Let $u_{\textrm{title}}$ and $u_{\textrm{desc}}$ be the number of hidden units of the title and description encoders obtained here.
The $n$ hidden states can be expressed as $H^{\textrm{title}} = \{\mathbf{h}^{\textrm{title}}_{1}, \cdots, \mathbf{h}^{\textrm{title}}_{n}\}$ and $H^{\textrm{desc}} = \{\mathbf{h}^{\textrm{desc}}_{1}, \cdots, 
\mathbf{h}^{\textrm{desc}}_{n}\}$, respectively.
Compute a vector $\mathbf{x}_{\textrm{feats}}$ that concatenates these hidden states, $H^{\textrm{title}}$, $H^{\textrm{desc}}$, one-hot vectors of gender features $\mathbf{x}_{\textrm{gender}} \in \mathbb{R}^{d_{\textrm{gender}}}$, and genre features $\mathbf{x}_{\textrm{genre}} \in \mathbb{R}^{d_{\textrm{genre}}}$:
\begin{eqnarray}
    \mathbf{x}_{\textrm{feats}} = \textrm{concat}(H^{\textrm{title}}, H^{\textrm{desc}}, \textbf{x}_{\textrm{genre}}, \textbf{x}_{\textrm{gender}}).
\end{eqnarray}
Note, $\mathbf{x}_{\textrm{feats}} \in \mathbb{R}^{d_{\textrm{feats}}}$;$d_{\textrm{feats}} = n \times (u_{\textrm{title}} + u_{\textrm{desc}}) + d_{\textrm{gender}} + d_{\textrm{genre}}$.
These concatenated vectors are inputted in a multi-layer perceptron (MLP) which is an output layer of the proposed framework.
To predict conversions $\hat{y}^{\textrm{(cv)}}$ and clicks $\hat{y}^{\textrm{(click)}}$, multi-task learning described later predicted $\hat{\mathbf{y}}_{\textrm{multi}} = \{\hat{y}^{\textrm{(cv)}}, \hat{y}^{\textrm{(click)}} \}$ through the MLP:
\begin{eqnarray}
    \hat{\mathbf{y}}_{\textrm{multi}} = \textrm{MLP}(\mathbf{x}_{\textrm{feats}}).
\end{eqnarray}
To improve the performance of the model robustness, we use wildcard training~\cite{shimada2016document} with dropout~\cite{hinton2012improving} for the input word embeddings.

\subsection{Multi-task Learning}
Conversion prediction is difficult, due to the imbalanced data, so we use the strategy of multi-task learning.
Multi-task learning is a method that solves multiple tasks related to each other, and that improves the prediction performance by learning common feature representations.
We adapt multi-task learning, and predict clicks and conversions prediction simultaneously.
Because click prediction may be related to conversion prediction, we expect to improve the prediction performance by learning common feature representations using multi-task learning.

In multi-task learning, the input is a feature vector of a training sample denoted by $\mathbf{x}$, and the ground truth is $y$.
For training samples $\mathbf{x} = \{\mathbf{x}_1, \mathbf{x}_2, \cdots, \mathbf{x}_N\}$, a single model, $f$, learns to generate predictions $\hat{y} = \{\hat{y}_1, \hat{y}_2, \cdots, \hat{y}_N\}$:
\begin{eqnarray}
    \hat{y} = f(\mathbf{x}_1, \mathbf{x}_2, \cdots, \mathbf{x}_N).
\end{eqnarray}
We minimize the mean squared error (MSE) over all samples, $N$, in $l = \frac{1}{N} \sum_{i=1}^{N} (y_i - \hat{y}_i)^2$.
In $K$ supervised tasks, the multi-task model, $F = \{f_1, f_2, \cdots, f_K\}$, learns to generate predictions $\hat{\mathbf{y}} = \{\hat{y}^{(1)}, \hat{y}^{(2)}, \cdots, \hat{y}^{(K)}\}$:
\begin{eqnarray}
     \hat{\mathbf{y}} = F(\mathbf{x}_{1}, \mathbf{x}_{2}, \cdots, \mathbf{x}_{N}).
\end{eqnarray}
The total loss is calculated from the sum of loss in each task,
\begin{eqnarray}
    \mathcal{L} = \frac{1}{N} \sum_{k=1}^{K} \sum_{i=1}^{N} \left(y_{i}^{(k)} - \hat{y}_{i}^{(k)}\right)^2.
\end{eqnarray}
In this task, for ground truth of $y^{\textrm{(cv)}}$ and $y^{\textrm{(click)}}$, we minimize losses for predicted conversions $\hat{y}^{\textrm{(cv)}}$ and clicks $\hat{y}^{\textrm{(click)}}$:
\begin{eqnarray}
    \mathcal{L}_{\textrm{multi}} = \frac{1}{N} \sum_{i=1}^{N} \left(y_{i}^{\textrm{(cv)}} - \hat{y}_{i}^{\textrm{(cv)}}\right)^2 + \lambda \frac{1}{N} \sum_{i=1}^{N} \left(y_{i}^{\textrm{(click)}} - \hat{y}_{i}^{\textrm{(click)}}\right)^2,
\end{eqnarray}
where $\lambda > 0$ is the hyper-parameter to control the effect of the click loss.

\begin{figure}[tb]
    \centering
    \includegraphics[width=0.9\linewidth]{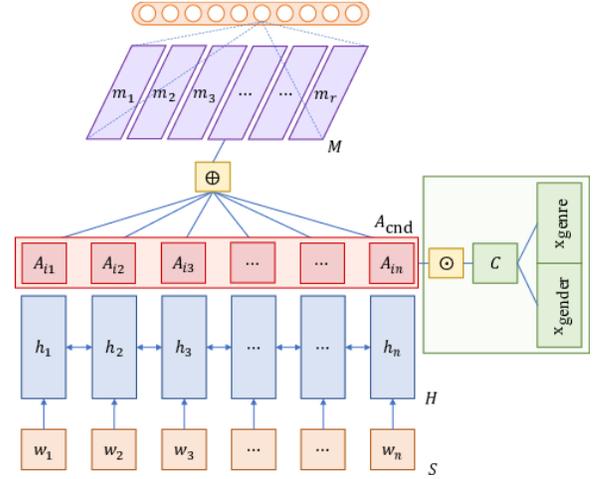}
    \caption{
        Example of the conditional attention mechanism. Conditional attention is calculated from the element-wise product of the attention matrix $A$ and the feature vector $\textbf{c}$ consisting of the gender and the genre.
    }
    \label{fig:example_of_conditional_attention}
\end{figure}

\subsection{Conditional Attention}
We propose the strategy of the conditional attention mechanism.
Supporting the creation of ad creatives by considering attribute values is useful, but the conventional attention mechanism learns keywords or key phrases, by calculating the alignment score using only the input sentence.

In this paper, we propose a conditional attention mechanism to calculate self-attention, using feature vectors obtained from the attribute values of the ad creative.
\figref{fig:example_of_conditional_attention} illustrates the conditional attention mechanism.
It can consider ad creative attributes against the conventional attention mechanism.

The conditional attention mechanism is calculated from the attention of the \textit{text encoder} and the feature vector obtained from the attribute values of the ad creative text.
Each word in the word sequence $S$ is independent of the others.
To capture these word order relations, we apply a \textit{text encoder} to the text, to obtain the hidden state $\textbf{h}_t \in \mathbb{R}^{u}$.
The $n$ hidden states of these $u~\times~n$ dimensions can be expressed as $H = \{\mathbf{h}_1, \mathbf{h}_2, \cdots, \mathbf{h}_n\}$.

To consider ad attribute values, a \textit{conditional} vector, $\textbf{c} \in \mathbb{R}^{n}$, is calculated by performing a linear combination of $\textbf{x}_{\rm feats} \in \mathbb{R}^{d_{\textrm{feats}}}$ and trainable parameters $W_{\textrm{prj}} \in \mathbb{R}^{n \times d_{\rm feats}}$:
\begin{eqnarray}
    {\bf c} = W_{\rm prj} \textbf{x}_{\rm feats}.
\end{eqnarray}
Here, we use \textit{self-attention}~\cite{lin2017structured} for computing the linear combination.
The attention mechanism takes the entire hidden state $H$ of the \textit{text encoder} as the input and outputs attention vector $\textbf{a}$:
\begin{eqnarray}
    \textbf{a} = \textrm{softmax}(\textbf{w}_{s2}^T\textrm{tanh}(W_{s1} H)),
\end{eqnarray}
where $W_{s1} \in \mathbb{R}^{n \times u}$ and $\textbf{w}_{s2} \in \mathbb{R}^{n}$ are trainable parameters.
Because $H$ is an $n \times u$ dimension, the size of attention vector $\textbf{a}$ is $n$.
The $\textrm{softmax}(\cdot)$ is calculated so that the sum of all the weight is 1.

Furthermore, we calculate the \textit{conditional attention vector} using the attributes given to the ad creative.
The \textit{conditional attention} vector, $\textbf{a}_{\textrm{cnd}}$, is calculated using conditional vector $\textbf{c}$ and attention vector $\textbf{a}$:
\begin{eqnarray}
    \textbf{a}_{\textrm{cnd}} = \textbf{a} \odot \textbf{c}.
\end{eqnarray}
Here, $\odot$ is an element-wise product.
We want $r$ different parts to be extracted from the ad creative texts. 
Thus, the \textit{conditional attention} vector $\textbf{a}_{\textrm{cnd}}$ becomes \textit{conditional attention} matrix $A_{\textrm{cnd}} \in \mathbb{R}^{n \times r}$.
Therefore, sentence vector $\mathbf{m}$ with the embedded ad creative text becomes sentence matrix $M \in \mathbb{R}^{u \times r}$.
The \textit{conditional attention} matrix, $A_{\textrm{cnd}}$, is multiplied by hidden state $H$ of the \textit{text encoder}, and the $r$-weighted sentence matrices are calculated as follows:
\begin{eqnarray}
    M = HA_{\rm cnd}.
\end{eqnarray}
In the proposed framework, the model makes predictions based on the calculated $M$ and ad creative attributes, such as $\textbf{x}_{\textrm{gender}}$ and $\textbf{x}_{\textrm{genre}}$.

\section{Experiments}\label{sec:experiment}
\begin{table}[tb]
\centering
\caption{
    Features included in the ad creative dataset.
    It contains 1,694 campaigns, some of which were part of campaigns delivered by Gunosy.
    The average lengths of the title and description texts are about 15 and, 32 characters, respectively.
    The Campaign ID feature is not directly inputted in the model, because the ID is used for evaluations with cross-validation based on the ID.
}
\label{tab:collect_features}
\resizebox{\columnwidth}{!}{%
\begin{tabular}{@{}c|lll@{}}
\toprule
\multicolumn{2}{c}{Features}                     & Feature Description      & Details               \\ \midrule \midrule
\multicolumn{2}{c}{Campaign ID}                  & Campaign ID in Gunosy Ads & 1,694 campaigns        \\ \midrule \midrule
\multirow{2}{*}{\rotatebox[origin=c]{90}{Texts}} & Title       & Title texts               & Avg. 15.44±3.16 chars \\
                                                 & Description & Description texts         & Avg. 32.69±5.43 chars \\ \midrule
\multirow{2}{*}{\rotatebox[origin=c]{90}{Attrs}} & Genre       & Genre of the creatives    & 20 types              \\
                                                 & Gender      & Gender of delivery target & 3 types               \\ \bottomrule
\end{tabular}%
}
\end{table}

\subsection{Dataset}
We use real-world data from the Japanese digital advertising program Gunosy Ads\footnote{\url{https://gunosy.co.jp/ad/}}, provided by Gunosy Inc.\footnote{\url{https://gunosy.co.jp/en/}}.
Gunosy Inc. is a provider of several news delivery applications, and Gunosy Ads delivers digital advertisements for these applications.
Gunosy is a news delivery application that achieved more than 24 million downloads in January 2019.

\begin{table*}[tb]
\centering
\caption{
    Comparison of the prediction performance of CVs in mean squared error (MSE) criteria.
    The proposed multi-task learning and conditional attention reduced MSE in almost all the categories, especially estimating cases where the number of conversions (\#CV) is one or more ($\#\textrm{CV} > 0$).
    However, ``All predicted as zero'' showed sufficiently low MSE in this category, due to too many $\textrm{CV} = 0$ in this dataset. Therefore, we conclude using MSE as an evaluation metric is not suitable in this study.
}
\label{tab:result_of_MSE}
\begin{tabular}{@{}clcccc@{}}
\toprule
\multicolumn{2}{c}{\multirow{3}{*}{Model}}                                                            & \multicolumn{4}{c}{MSE}                                  \\ \cmidrule(l){3-6} 
\multicolumn{2}{c}{}                                                                                  & \multicolumn{2}{c}{All}             & \multicolumn{2}{c}{\#CV \textgreater 0} \\ \cmidrule(l){3-6} 
\multicolumn{2}{c}{}                                                                                  & Single-task      & \textbf{Multi-task}       & Single-task        & \textbf{Multi-task}         \\ \cmidrule(r){1-2} \cmidrule(lr){3-4} \cmidrule(l){5-6}
\multicolumn{2}{c}{MLP}                                                                               & 0.01712          & 0.01698          & 0.04735            & 0.03199            \\ \cmidrule(r){1-2} \cmidrule(lr){3-4} \cmidrule(l){5-6}
\multirow{3}{*}{GRU}                                                          & Vanilla               & 0.01696          & 0.01695          & 0.04657            & 0.04355            \\
                                                                              & Attention             & 0.01685          & 0.01688          & 0.04695            & 0.03105            \\
                                                                              & \textbf{Conditional attention} & \textbf{0.01683} & \textbf{0.01675} & \textbf{0.04641}   & \textbf{0.02825}   \\ \cmidrule(r){1-2} \cmidrule(lr){3-4} \cmidrule(l){5-6}
\multicolumn{2}{c}{\textbf{All predicted as zero}}                                                          & \multicolumn{2}{c}{\textbf{0.02148}} & \multicolumn{2}{c}{---}                                          \\ \bottomrule
\end{tabular}%
\end{table*}

For evaluation, we used 14,000 ad creatives delivered by Gunosy Ads from August 2017 to August 2018.
In digital advertising, the cost of acquiring a conversion is called the cost per acquisition (CPA).
Advertisers set target CPAs for a product, and manage its ad creatives to improve their performance.
When the target CPAs for creatives are different, the trend of conversions may also vary, and for this reason, the dataset we selected comprises ad creatives where the target CPA was within a certain range.
In addition, we removed creatives with a low number of impressions\footnote{An occasion when a particular advertisement is seen by someone using the application.} from the dataset.
As shown in \tabref{tab:collect_features}, the title, description, and genre of the ad creative, as well as the gender to which the ad is delivered, are used as input features.
Note that the Campaign ID is not a feature directly used as an input in the model, because the ID is used for evaluating with cross-validation based on the ID.

Creative texts written in Japanese are split into words using MeCab~\cite{kudo2006mecab}, a morphological analysis engine for Japanese texts, and mecab-ipadic-neologd~\cite{sato2015mecabipadicneologd}, which is a customized system dictionary that includes many neologisms for MeCab.
The number of clicks and conversions is log-normalized.
\begin{figure}[tb]
    \centering
    \includegraphics[width=0.65\linewidth]{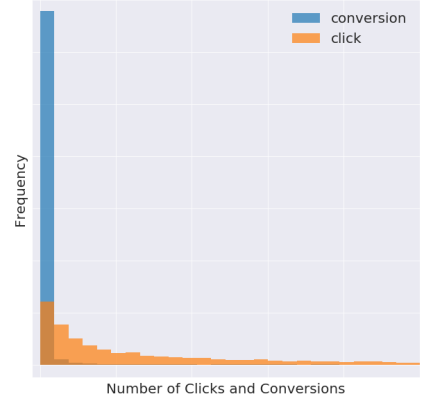}
    \caption{
        Distribution of clicks and conversions in the dataset.
        The number of conversions is concentrated on zero. Compared with the number of conversions, the number of clicks indicates a long-tail distribution.
    }
    \label{fig:hist_of_click_conversion}
\end{figure}

\begin{figure}[tb]
    \centering
    \includegraphics[width=0.65\linewidth]{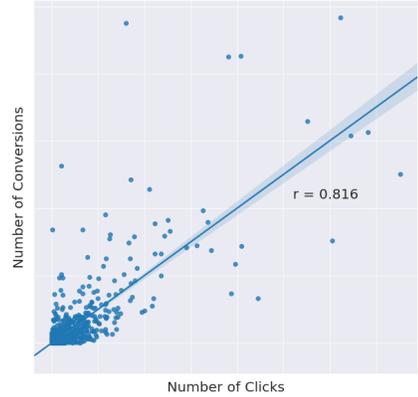}
    \caption{
    The linear relation between the clicks and conversions in the dataset (correlation coefficient $r=0.816$).}
    \label{fig:corr_click_and_cv}
\end{figure}

\figref{fig:hist_of_click_conversion} shows a histogram of the number of clicks and conversions.
The number of conversions is concentrated on zero, and in relation, the number of clicks is a long-tailed distribution.
Therefore, the ad creative dataset is definitely imbalanced.
\figref{fig:corr_click_and_cv} shows the distribution between the number of clicks and conversions in the dataset.
The correlation coefficient between the number of clicks and conversions is 0.816, which is a strong correlation. As a reminder, we hide the number of clicks and conversions, also their frequencies, for confidentiality reasons.

\subsection{Experimental Settings}
In these experiments, support vector regression (SVR) and an MLP-based \textit{text encoder} were used as a baseline model.
When inputting creative text in the SVR model, we used average-pooled sentence representations computed from word representations, using pre-trained word2vec (w2v)~\cite{suzuki2018joint}.
The same pre-trained w2v was used as word embedding for the proposed model.

We compared and examined the following models: MLP (not considering word order) and GRU (considering word order) as the \textit{text encoder} in the proposed framework.
LSTM was also considered as a candidate for the baseline model; however, it showed no improvement in performance, so it was excluded from the experiment.
In addition, CNNs are known to be capable of training at high speed, because they can perform parallel calculations, compared with LSTM and GRU, and their performances are also known to be equal. 
Nevertheless, these methods were excluded in these experiments, because we were targeting an RNN-based model that can apply attention for visualizing the contributions of words to ad creative evaluation.

We compared the proposed models used in the proposed framework.
The following models were compared and examined, to confirm the effect of multi-task learning in conversion prediction:
\begin{description}
    \item[Single-task:] A commonly known model that predicts conversions only; and
    \item[Multi-task:] A model that simultaneously predicts the number of clicks and the number of conversions.
\end{description}

To confirm the effect of the conditional attention mechanism, we compared the following models:
\begin{description}
    \item[Vanilla:] A simple \textit{text encoder} without an attention mechanism. It is a baseline in the proposed model;
    \item[Attention:] A mechanism that introduces self-attention to the \textit{text encoder}. It makes it possible to visualize which word contributed to prediction during creative evaluation; and
    \item[Conditional Attention:] A mechanism introduced to the \textit{text encoder} of the proposed method. Conditional attention can be computed and visualized considering the attribute values of the ad creative. Different attentions can be visualized by changing the attribute value for the same creative text.
\end{description}

In addition, the hyper-parameter setting is described below.
The mini-batch size was set to be 64, and the number of epochs was set to be 50.
For multi-task learning, we used a fixed value of $\lambda = 1$.
In the \textit{text~encoder}, the number of hidden units was set to be 200 for $u_{\textrm{title}}$ and $u_{\textrm{desc}}$.
For all models, we use Adam \cite{kingma2014adam}, with a weight decay of $1\mathrm{e}^{-4}$, for parameter optimization.

\subsection{Evaluation Metrics}
First, as evaluation metrics, we adopt not only MSE but also normalized discounted cumulative gain (NDCG)~\cite{jarvelin2002cumulated}, which is evaluation metrics for ranking.
MSE measures the average of the squares of the errors, which is the average squared difference between the estimated values and what is estimated. 
We adopted ranking evaluation metrics because the number of conversions is imbalanced.
As shown in \figref{fig:hist_of_click_conversion}, most ad creative conversions are zero and imbalanced. A high evaluation score can be achieved by an overfit model that predicts all outputs as zero when such metrics are used.
For the creation of high-performing ad creatives, rather than predicting zero conversions, we would like to accurately predict high-conversion creatives as such.

\begin{table*}[tb]
\centering
\caption{
    Comparison of the normalized discounted cumulative gain (NDCG) in the proposed model.
    When calculating NDCG scores, the results for all data and the scores restricted to the top 1\% of conversions (\#CV) were calculated.
}
\label{tab:result_of_NDCG}
\begin{tabular}{@{}clcccc@{}}
\toprule
\multicolumn{2}{c}{\multirow{3}{*}{Model}}                                                            & \multicolumn{4}{c}{NDCG {[}\%{]}}                                   \\ \cmidrule(l){3-6} 
\multicolumn{2}{c}{}                                                                                  & \multicolumn{2}{c}{All}         & \multicolumn{2}{c}{\#CV top 1 \%} \\ \cmidrule(l){3-6} 
\multicolumn{2}{c}{}                                                                                  & single         & \textbf{multi-task}     & single          & \textbf{multi-task}      \\ \cmidrule(r){1-2} \cmidrule(lr){3-4} \cmidrule(l){5-6}
\multicolumn{2}{c}{SVM}                                                                               & \multicolumn{2}{c}{96.72}       & \multicolumn{2}{c}{83.73}         \\ \cmidrule(r){1-2} \cmidrule(lr){3-4} \cmidrule(l){5-6}
\multicolumn{2}{c}{MLP}                                                                               & 96.68          & 97.18          & 82.97           & 84.12           \\ \cmidrule(r){1-2} \cmidrule(lr){3-4} \cmidrule(l){5-6}
\multirow{3}{*}{GRU}                                                          & Vanilla               & 96.54          & 97.00          & 76.39           & 78.51           \\
                                                                              & Attention             & 96.76          & 97.11          & 83.00           & 85.49           \\
                                                                              & \textbf{Conditional Attention} & \textbf{96.77} & \textbf{97.20} & \textbf{87.11}  & \textbf{87.14}  \\ \bottomrule
\end{tabular}%
\end{table*}

NDCG is mainly used in the experiments.
NDCG is the discounted cumulative gain (DCG) normalized score.
In DCG, the score decreases as the evaluation of an advertisement declines, so a penalty is imposed if a low effect is predicted for highly effective creatives.
At the time of the NDCG calculation, after obtaining the rank of the ground truth, and its predicted value, respectively, evaluation scores are calculated for all the evaluation data, as well as those restricted to the top 1\% of conversions.

For ad creative evaluation, the metrics are computed with cross-validation.
In most advertising systems, advertisements are delivered in units of campaigns.
In a campaign, the target gender and its genre are set, and multiple ad creatives are developed.

In this paper, we predict the number of conversions for ad creative text in unknown campaigns, and confirm the generalization performance of the proposed framework.
Therefore, at the time of the evaluation, five-fold cross-validation was performed in such a manner that the delivered campaigns did not overlap.

\vspace{-2mm}
\subsection{Experimental Results}
For confirming the accuracy of the proposed framework compared with the baselines, we compared single-task and multi-task learning, and the results of the application of the conditional attention mechanism are described.
Through almost all the results, the proposed framework applying multi-task learning and the conditional attention mechanism achieved a better performance than the other methods.
Especially, when focusing on ad creatives with many conversions, the proposed framework achieved high prediction accuracy.

\tabref{tab:result_of_MSE} shows the MSE score with all the evaluations in each model, and with one or more conversions in each model.
Almost all the results show that the model applying the multi-task learning and conditional attention mechanism had a smaller MSE score than the other models did.
Overall, the RNN-based GRU showed better performance than the baseline models.
Therefore, the results suggest that it is important to properly capture word order when evaluating creative texts.
Compared with \textit{vanilla} and \textit{attention}, in the proposed model, \textit{conditional attention} showed a better performance.

Although the improvement of all datasets is weak, because as shown in \figref{fig:hist_of_click_conversion}, the number of conversions of many ad creatives is zero, the MSE is small, even if the conversion of most ad creatives is predicted to be zero.
Therefore, we evaluated data with conversions other than zero.
As a result, we found that the proposed model exhibits much better performance than the baseline model for data with one or more conversions.
The proposed model was able to predict creatives with more conversions than the baseline models.

To evaluate ad creatives with many conversions as such, we used the ranking algorithm NDCG.
The NDCG result in the proposed model is shown in \tabref{tab:result_of_NDCG}\footnote{The same tendency was observed even when mean average precision (MAP) was used as an evaluation metric.}.
The NDCG score (regarded as \textit{All} in \tabref{tab:result_of_NDCG}) for all the datasets is shown for reference, because as noted above, most samples have zero conversions.
The performance of the GRU model that considers word order compared with the baseline model improved by an average of approximately 3-5\%, with many conversions.

In the NDCG result (\tabref{tab:result_of_NDCG}), the multi-task model realized higher prediction accuracy than the single-task model predicting only conversions did.
A score improvement of approximately 1-2\% was confirmed when compared with the baselines.
Because clicks are highly correlated with target ad conversions, as shown in \figref{fig:corr_click_and_cv}, rather than predicting conversions alone, training the model to multi-task by predicting clicks simultaneously can improve prediction accuracy.
By training clicks and conversions, the proposed model seems to implicitly learn features that contribute to conversion prediction.

\begin{table}[tb]
\centering
\caption{
    Comparison of NDCG between the CVR directly predicted by the single-task model and the CVR (\#conversions / \#clicks) calculated from the multi-task GRU model's predicted conversions and clicks. 
    The threshold value for calculating NDCG is assumed to be a CVR of 0.5 or more.
}
\label{tab:result_clickcv_vs_cvr}
\begin{tabular}{@{}clc@{}}
\toprule
\multicolumn{2}{c}{Model}                                             & NDCG {[}\%{]}  \\ \cmidrule(r){1-2} \cmidrule(l){3-3}
\multirow{3}{*}{Single-task}         & Vanilla                        & 80.54          \\
                                     & Attention                      & 82.58          \\
                                     & \textbf{Conditional attention} & 83.89          \\ \cmidrule(r){1-2} \cmidrule(l){3-3}
\multirow{3}{*}{\textbf{Multi-task}} & Vanilla                        & 82.63          \\
                                     & Attention                      & 84.27          \\
                                     & \textbf{Conditional attention} & \textbf{85.61} \\ \bottomrule
\end{tabular}
\end{table}
Because several previous studies predicted the CVR directly, we also calculated it, using the prediction of the multi-task learning model, and compared the accuracy.
In a multi-task model, the CVR can be calculated by dividing conversions by clicks.
In \tabref{tab:result_clickcv_vs_cvr}, the multi-task model is compared with the single-task model by directly estimating the CVR.
The prediction performance of the multi-task model is higher than that of the single-task model.
Although the number of clicks and conversions predicted by multi-task learning may not always be close to the ground truth, the ratio of the number of clicks to the conversion number is captured properly.

\begin{table}[tb]
\centering
\caption{
    Comparison of GRU models for creative texts and their attribute value interactions.
    Performance is improved using conditional attention rather than giving attribute values directly to word vectors.
}
\label{tab:result_word_embedding_attribute}
\resizebox{\columnwidth}{!}{%
\begin{tabular}{@{}cccc@{}}
\toprule
\multicolumn{2}{c}{\multirow{2}{*}{Model}}                         & \multicolumn{2}{c}{NDCG {[}\%{]}}    \\ \cmidrule(l){3-4} 
\multicolumn{2}{c}{}                                               & Single-task    & \textbf{Multi-task} \\ \cmidrule(r){1-2} \cmidrule(l){3-4}
\multirow{2}{*}{w2v + attributes} & Vanilla                        & 77.84          & 78.03               \\
                                  & Attention                      & 80.39          & 83.52               \\ \cmidrule(r){1-1} \cmidrule(lr){2-2} \cmidrule(l){3-4}
\textbf{w2v}                      & \textbf{Conditional attention} & \textbf{87.11} & \textbf{87.14}      \\ \bottomrule
\end{tabular}%
}
\end{table}

In \tabref{tab:result_of_NDCG}, the conditional attention mechanism achieved better results the NDCG metric.
In particular, the conditional attention mechanism showed better results than the conventional attention mechanism did.
In the conventional attention mechanism, the training was focused solely on the co-occurrence relation between words in the input text, but the conditional attention mechanism can predict conversion by using the attribute value.

\tabref{tab:result_word_embedding_attribute} shows the result comparing feature interaction between w2v-based embeddings and ad attribute values.
In the proposed framework, this interaction is realized with the conditional attention mechanism, explicitly.
Because attention is computed by the input variables, this interaction is implicitly expressed by inputting both variables in the \textit{text encoder}.
For confirming the effect of this explicit interaction in the conditional attention mechanism, we compared the model that inputted both variables in the \textit{text encoder} with the conditional attention mechanism.
The conditional attention mechanism showed the best performance in the single-task and multi-task model.
Introducing the vanilla model and the conventional attention model to the word representation with ad attribute values resulted in a poor performance, mainly because the duplicate interactions were calculated excessively.
It is suggested that it is better to introduce the explicit interaction of attribute values.

\section{Discussion}\label{sec:discussion}
    \subsection{Advantages of the Proposed Framework}
The proposed framework aimed to predict not the CVR but conversions.
However, in CVR prediction, we also achieved high performance using multi-task learning results.
From the business perspective, we assume that predicting conversions can evaluate high-performing ad creatives, rather than predicting the CVR.
In the process of advertising management, advertisers stop low-performing creatives and focus cost on high-performing creatives, so there are few conversions of low-performing creatives, and many conversions of high-performing creatives.
For that reason, the number of conversions seems to be a good metric for evaluating ad creatives, and conversion prediction may be learn good representation of high-performing ad creatives.

We proposed an RNN-based framework, and achieved high-performance conversion prediction.
Normally, when advertisers create the creative text, words are selected in such a way as to change the word order or emphasize the characteristics of the product.
We let the model learn feature representation so that it could properly capture the features between words in creative text.

\begin{figure}[tb]
    \centering
    \subfloat[
    Title text: "Chosen by 10 million people! The 10 games played by everyone."
    Description text: "Exclusively introducing free games that you will want to install on mobile phone."
    ]{
        \includegraphics[clip, width=0.9\linewidth]{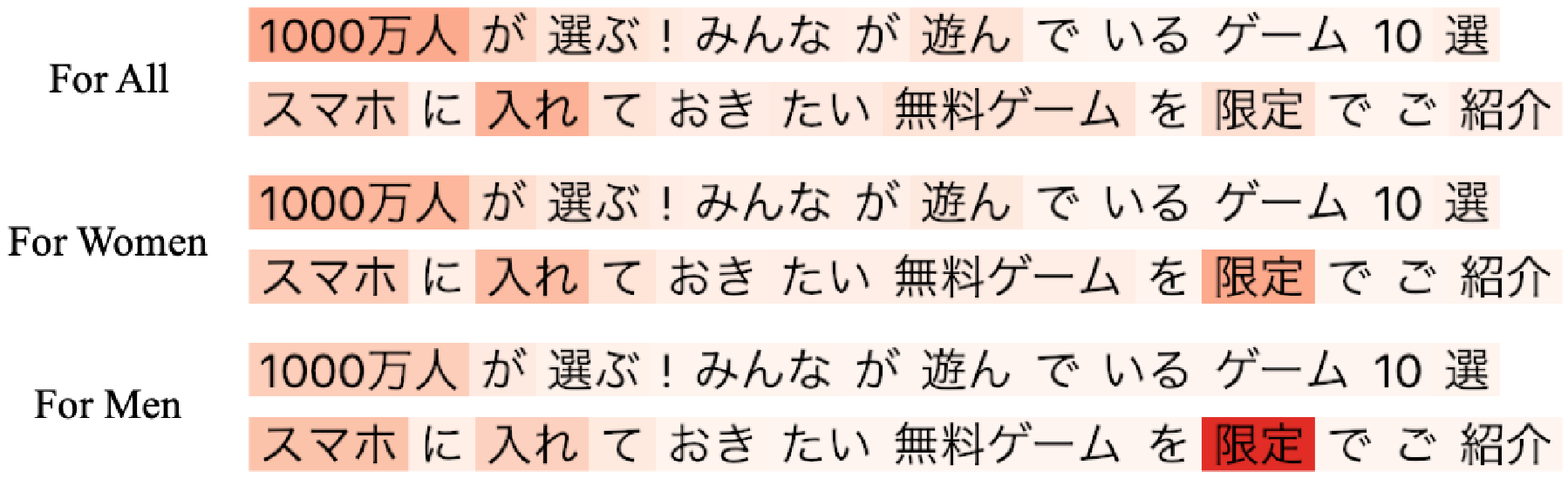}
        \label{fig:label-A}
    } \\
    \subfloat[
    Title text: "Success in -10 kg weight loss! This is the reason for getting slim."
    Description text:  "Realizing the effects popular among girls."
    ]{
        \includegraphics[clip, width=0.9\linewidth]{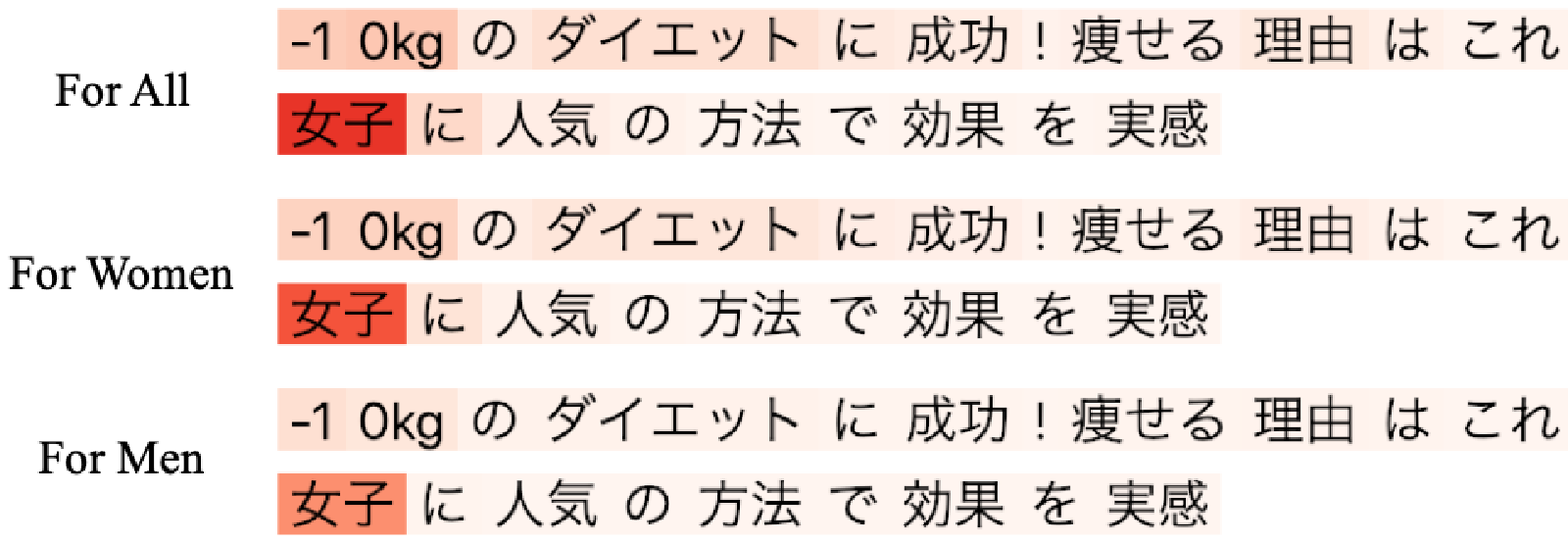}
        \label{fig:label-B}
    } \\
    \subfloat[
    Title text: "Supervised by a famous celebrity; easy cookbook."
    Description text: "Recommended for men living alone!"
    ]{
        \includegraphics[clip, width=0.9\linewidth]{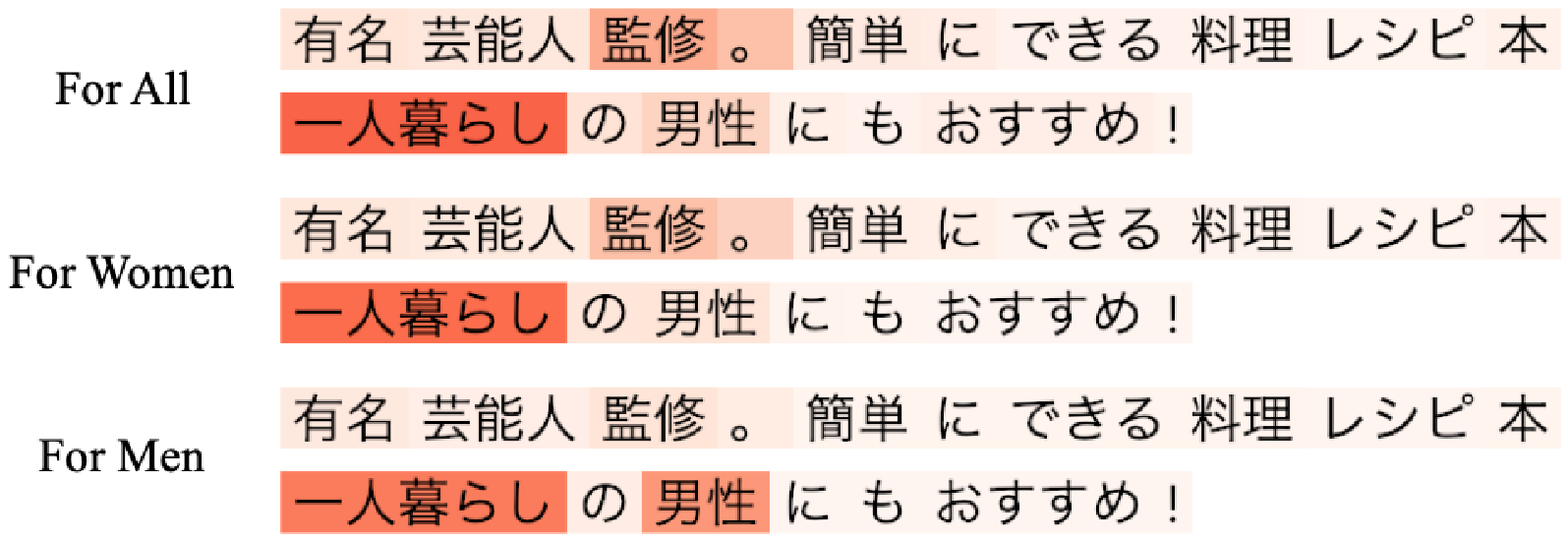}
        \label{fig:label-C}
    }
    \caption{Heatmap showing the change in conditional attention when the distribution target is changed.}
    \label{fig:visualize_conditional_attentions}
\end{figure}

We achieved high-performance conversion prediction by predicting the clicks and conversions simultaneously; this method is called multi-task learning.
Many ad creative conversions are zero, which is imbalanced data, so predicting this number correctly is a difficult task.
Multi-task learning is a method that learns multiple tasks related to each other, and improves prediction performance.
Because ad click actions represent the pre-action of conversion actions, we assumed that click prediction may be related to conversion prediction, and that improved conversion prediction would be obtained using multi-task learning.
We expect that this achievement can be applied to various prediction tasks with imbalanced data.

High accuracy was achieved by conditional attention in the experiment.
When predicting the CTR or CVR of advertisements, it is important to properly capture the explicit feature interactions~\cite{lian2018xdeepfm}.
The conditional attention mechanism seems to capture the explicit interactions between the attention gained from creative text and feature representations consisting of the text's attribute values.
It is also possible to visualize different forms of attention by controlling different attribute values in the same creative texts. 
This can greatly support ad creative creation.

\subsection{Visualization for High-Performance Ad Creative Creation}
We attempt to highlight important words using attention.
If the words contributing to conversions are clarified, advertisers will be able to easily create high-performing ad creatives.
Attention is a mechanism that focuses on words contributing to prediction, and the results predicted by these mechanisms are useful for creating ad creatives.
The proposed conditional attention mechanism can compute attention based on ad creative attributes, as well as the genre and target gender, so conditional attention highlights important words according to their attribute values.

\figref{fig:visualize_conditional_attentions} shows examples of the visualization of attention when modifying the attributes of gender for three Japanese ad creative texts for different groups (for all audiences, for women, and for men).
Different types of attention were gained using conditional attention mechanism.

\begin{CJK*}{UTF8}{min}
\figref{fig:label-A} shows an ad creative for a mobile game.
The word ``1000万'' (\textit{10 million}), a concrete numerical value, and the word ``限定'' (\textit{exclusively}) contribute to predicting conversion.
Especially for men, the word ``限定'' contributes more to the prediction than it does for women.

\figref{fig:label-B} is an ad creative in the beauty genre for women.
The word ``女性'' (\textit{girls}) contributes to the conversion prediction.
More attention is also given to ``ダイエット'' (\textit{weight loss}) for women than men.
When setting the delivery target to men in this ad creative, the attention score and the number of predicted conversions are smaller than that of all targets or female targets.

\figref{fig:label-C} is an ad creative in the health food genre for men.
The words ``一人暮らし'' (\textit{living alone}) and ``監修'' (\textit{supervised by}) are closely highlighted.
The word ``lived alone'' is an expression that narrows down the delivery target.
When proposing ad creative text, the term ``supervised by'' is often used in conjunction with the names of celebrities, and the effect is high.
Moreover, it was confirmed that the word ``男性'' (\textit{men}) is an important factor when the delivery target is male.
\end{CJK*}

Overall, most words that are highlighted by attention are concrete numerical values and expressions focusing on the delivery target.
We believe that this knowledge is also empirically correct.
In this way, visualization of important words using the conditional attention mechanism of the proposed method can be expected to greatly contribute to supporting the creation of ad creatives.
This result is a good example of interpretability.        

\vspace{-1.0mm}
\vspace{-1.0mm}
\section{Conclusion}\label{sec:conclusion}
In this paper, we propose a new framework to support the creation of high-performing ad creative text.
The proposed framework includes three key ideas, multi-task learning and conditional attention improve prediction performance of advertisement conversion, and attention highlighting offers important words and/or phrases in text creatives.
We confirmed that the proposed framework realizes an excellent performance thanks to these ideas, through experiments with actual delivery history data.

In the future, we will build a framework that simultaneously uses images attached to ad creatives, and aim to improve the accuracy of conversion prediction.
%
\bibliographystyle{ACM-Reference-Format}
\bibliography{reference}

\end{document}